\documentclass[lettersize,journal]{IEEEtran}

\usepackage{amsmath,amsfonts}
\usepackage{algorithmic}
\usepackage{algorithm}
\usepackage{array}
\usepackage{subcaption}
\usepackage{textcomp}
\usepackage{stfloats}
\usepackage{url}
\usepackage{verbatim}
\usepackage{graphicx}
\usepackage{cite}
\usepackage{hyperref}
\usepackage{booktabs}
\usepackage{nicefrac}
\usepackage{microtype}
\usepackage{bm}
\usepackage{tabularx}
\usepackage{marvosym}

\graphicspath{ {./img/} }

\title{AgilePE: Autonomous UAV Pursuit-Evasion via Self-Play Reinforcement Learning}

\author{Wenhao Tang$^{1*}$, Tianyang Chen$^{2*}$, Zhejun Cui$^{2*}$, Boyuan An$^{2}$, Jiayu Chen$^{1}$, Ruize Zhang$^{1}$, \\ Huidong Liu$^{3}$, Tianyue Wu$^{2}$, Qingmin Liao$^{1}$, Fei Gao$^{2}\textsuperscript{\Letter}$, Yu Wang$^{1}\textsuperscript{\Letter}$, Chao Yu$^{1}\textsuperscript{\Letter}$
\thanks{$^{*}$Equal Contributions.}
\thanks{{\Letter} Corresponding Authors.} 
\thanks{$^{1}$Tsinghua University, Beijing, 100084, China.}
\thanks{$^{2}$Zhejiang University, Hangzhou, 310058, China.}
\thanks{$^{3}$Chongqing University, Chongqing, 400044, China.}
}

\begin{document}

\maketitle

\begin{figure*}[h]
\centering
\includegraphics[width=0.8\textwidth]{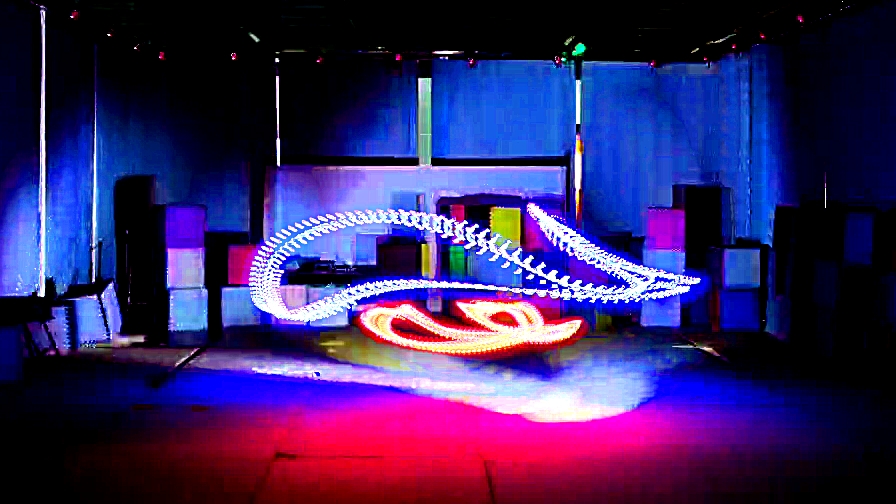}
\caption{A long-exposure photograph of a real-world pursuit-evasion experiment in a motion-capture arena. The red and blue light trails visualize the flight trajectories of the physical pursuer and evader, respectively, demonstrating agile maneuvers and emergent tactical behaviors during zero-shot sim-to-real deployment.}
\label{fig:teaser}
\end{figure*}

\begin{abstract}
Autonomous pursuit-evasion represents a fundamental challenge for Unmanned Aerial Vehicles (UAVs) operating in dynamic and adversarial environments, where agents must make rapid decisions under highly coupled motion constraints and continuously changing opponent behaviors. Traditional approaches typically rely on rule-based heuristics or differential game formulations, which often struggle to scale to high-dimensional aerial interactions and agile maneuvering scenarios.

In this paper, we present \textbf{AgilePE}, a complete system for autonomous UAV pursuit-evasion via self-play reinforcement learning. The proposed system jointly integrates agile low-level flight control, competitive multi-agent policy optimization, and sim-to-real deployment into a unified learning framework. At the control level, the policy directly maps onboard state observations to low-level Collective Thrust and Body Rates (CTBR) commands, enabling end-to-end agile maneuvering without intermediate trajectory planners or waypoint controllers. At the learning level, we employ competitive self-play reinforcement learning with Prioritized Fictitious Self-Play (PFSP) and a diversified opponent pool, allowing agents to continuously evolve through adversarial interactions with historical policies. This training paradigm stabilizes optimization, mitigates policy oscillation, and naturally leads to the emergence of sophisticated pursuit and evasion strategies. 

To support reliable real-world deployment, we further develop a hardware-aligned simulation pipeline incorporating calibrated actuator dynamics, communication latency modeling, and extensive domain randomization. The resulting policies can be transferred zero-shot from simulation to real quadrotor platforms without task-specific tuning. Extensive real-world flight experiments demonstrate that AgilePE achieves robust autonomous pursuit and evasion behaviors under sensor noise, aerodynamic disturbances, and dynamic environmental uncertainties. Emergent tactical behaviors, including rapid dodging and cooperative flanking maneuvers, further validate the effectiveness of the proposed system for real-world adversarial UAV autonomy.
\end{abstract}


\section{Introduction}

The operational scope of UAVs\cite{keane2013brief} has evolved dramatically over the past decades. Early research focused primarily on single-UAV autonomy, including pre-programmed navigation, terrain surveillance, and point-to-point trajectory tracking\cite{shakhatreh2019unmanned}. As onboard sensing and communication capabilities matured, the frontier shifted toward multi-UAV cooperation, enabling collaborative search and rescue\cite{lin2009uav}, distributed mapping\cite{colomina2014unmanned}, and formation flying\cite{chung2018survey}. More recently, attention has turned to adversarial multi-UAV operations, where autonomous systems contend with intelligent, non-cooperative opponents. Representative missions include the interception of hostile drones in restricted airspace\cite{moreira2019interception,stasinchuk2021multi,vrba2022autonomous}, autonomous aerial combat\cite{kong2020uav,li2025hierarchical,luo2025review}, and area-denial scenarios\cite{wang2021counter}. In these settings, UAVs must fly at their aerodynamic limits within highly non-stationary environments, making real-time autonomous decision-making essential. Removing human operators from the loop is necessitated by inherent constraints of telerobotics\cite{sheridan1992telerobotics}, including communication latency, signal jamming, and cognitive overload during high-speed engagements.

Among the diverse spectrum of multi-UAV adversarial tasks, the pursuit-evasion (PE) game\cite{isaacs1965differential,vidal2002probabilistic} stands out as one of the most representative paradigms. Conceptualized as a zero-sum competition\cite{von1944theory}, PE captures the fundamental antagonism inherent in aerial confrontation: pursuers seek to intercept evaders, while evaders attempt to delay or avoid capture indefinitely. Its practical relevance spans counter-Unmanned Aircraft System (counter-UAS) operations against unauthorized intruders\cite{wang2021counter}, close-range air-combat maneuvering\cite{yang2019maneuver}, and tactical interception in contested environments\cite{stasinchuk2021multi,vrba2022autonomous}. We choose PE as our research focus because it serves as a canonical benchmark for autonomous decision-making under antagonism while imposing stringent demands on flight agility: success depends on exploiting the full maneuvering envelope at sub-second timescales. Consequently, PE provides an ideal testbed for tightly integrating high-level tactical reasoning with low-level physical control.

The pursuit-evasion game has been tackled from three complementary angles. Classical differential-game theory, most notably Isaacs' Hamilton-Jacobi-Isaacs (HJI) framework\cite{isaacs1965differential,mitchell2005time}, yields optimal strategies for simple kinematics but falls victim to the curse of dimensionality when scaled to six-degree-of-freedom (6-DOF) aerial dynamics. Model Predictive Control (MPC)\cite{richalet1978model} offers a receding-horizon alternative, yet its efficacy depends on accurate predictive models of adversarial motion. More recently, Deep Reinforcement Learning (DRL)\cite{mnih2015human} has bypassed explicit analytical models and demonstrated promising results in simulated PE scenarios. Nevertheless, three critical limitations remain unresolved. \textbf{First}, most DRL frameworks output high-level kinematic commands, such as waypoints\cite{osborne2005waypoint} or velocity vectors, rather than direct actuator commands, constraining agility through intermediary planners. \textbf{Second}, adversarial training in competitive settings is notoriously unstable: policies oscillate and suffer from catastrophic forgetting\cite{mccloskey1989catastrophic} as the opponent distribution shifts continuously during co-evolution. \textbf{Third}, even when policies converge in simulation, the Sim-to-real gap\cite{rusu2016sim}, stemming from unmodeled aerodynamics, actuator latency, and sensor noise, frequently causes learned controllers to fail upon deployment on physical hardware.

To address these challenges, this study proposes \textbf{AgilePE}, a tightly integrated system for autonomous UAV PE. The framework comprises three stages, each targeting a specific gap in the literature.

\begin{itemize}
\item \textbf{End-to-End Agile Control Architecture.} At the perception-action layer, AgilePE replaces cascaded planners with an integrated end-to-end subsystem mapping state observations directly to low-level CTBR\cite{mellinger2011minimum}. This bypasses intermediary waypoint controllers and grants the neural network unrestricted access to the quadrotor's full flight envelope. A multi-objective reward framework balances adversarial goals with physical safety and actuation smoothness, ensuring convergence without hand-crafted heuristics.
\item \textbf{Bilateral Adversarial Training Pipeline.} At the algorithmic layer, we design a hierarchical training subsystem integrating bootstrapped warm-up, multi-dimensional curriculum learning\cite{bengio2009curriculum}, and competitive self-play from naive SP\cite{samuel1959some} to FSP\cite{heinrich2015fictitious} and PFSP\cite{vinyals2019grandmaster}. As pursuer and evader co-evolve, agents autonomously discover sophisticated tactics such as baiting, reactive dodging, and counter-pursuit, validating that advanced strategies can emerge purely through algorithmic evolutionary pressure.
\item \textbf{Hardware-Aligned Sim-to-Real Deployment Pipeline.} At the deployment layer, a high-fidelity kinematic backend with calibrated actuator models replicates thrust latency, rate delays, and multiplicative noise observed on physical hardware. Extensive domain randomization\cite{tobin2017domain} over mass, aerodynamics, and latency forces the policy to learn invariant control features. A two-stage reward refinement mechanism smooths control commands, enabling zero-shot deployment onto physical quadrotors as shown in Fig.~\ref{fig:teaser}.
\end{itemize}

The main contributions of this work are threefold. First, we propose an end-to-end agile control architecture that directly outputs CTBR commands, eliminating intermediary planners and unlocking the full flight envelope. Second, we introduce a bilateral adversarial training pipeline that progressively escalates from naive SP to FSP and PFSP, suppressing catastrophic forgetting and sustaining tactical diversity throughout co-evolution. Third, we present a hardware-aligned sim-to-real pipeline that enables zero-shot transfer to physical quadrotors without fine-tuning. Extensive simulation and real-world experiments validate robust policy equilibria and emergent tactical behaviors under zero-shot deployment.

\section{Related Work}

\subsection{Methodologies for UAV Pursuit-Evasion Games}

The PE game provides a rigorous mathematical framework for analyzing antagonistic aerial interactions, dating back to Isaacs' seminal work on differential games~\cite{isaacs1965differential}. Early approaches such as Artificial Potential Fields (APF)~\cite{khatib1986real} and MPC~\cite{richalet1978model} offered computationally efficient or constraint-aware solutions, yet they remain limited by local minima, susceptibility to unpredictable maneuvers, and reliance on accurate forward-predictive models. Differential game theory yields mathematically optimal strategies via HJI equations~\cite{isaacs1965differential,mitchell2005time}, but exact solutions suffer from the curse of dimensionality and become intractable for complex six-degree-of-freedom (6-DOF) UAV dynamics. More recently, DRL~\cite{mnih2015human} has emerged as a powerful paradigm for high-dimensional policy optimization. Nevertheless, existing DRL studies for aerial PE predominantly operate at high abstraction levels, such as velocity commands or waypoint planning~\cite{kong2020uav,li2025hierarchical,vrba2022autonomous}, whereas peak physical agility demands extending DRL to low-level actuator commands, particularly CTBR, a modality shown to significantly outperform velocity-level policies in agile flight benchmarks~\cite{kaufmann2022benchmark}.

\subsection{Adversarial Learning and Self-Play in UAV Tasks}

Training UAV policies against static or hand-scripted opponents typically yields brittle behaviors that generalize poorly to novel adversarial tactics. In physically grounded aerial tasks, ranging from air combat to multi-drone sports, self-play (SP) has become the dominant paradigm for generating an automatic curriculum of progressively stronger opponents~\cite{samuel1959some}. Nevertheless, naive SP is vulnerable to strategy cycling and unilateral collapse~\cite{zhang2024survey}. To stabilize co-evolution, Fictitious Self-Play (FSP) maintains a historical policy pool~\cite{heinrich2016deep}, while Prioritized Fictitious Self-Play (PFSP) concentrates training on the most challenging opponents via a win-rate matrix~\cite{vinyals2019grandmaster}. Recent hierarchical architectures such as HCSP further decouple high-level strategy from low-level motor control in multi-drone volleyball tasks, eliciting emergent team behaviors such as role switching~\cite{xu2025volleybots,zhang2025mastering}. Despite these advances, applying hierarchical adversarial training to PE remains largely unexplored, particularly when policies must output low-level actuator commands and transfer zero-shot to physical hardware.

\subsection{Sim-to-Real Transfer for Agile Flight}

The reality gap between simulation and physical deployment stems from unmodeled aerodynamic effects, sensor noise, latency, and parameter mismatch. Successful zero-shot transfer typically combines System Identification (SysID) with Domain Randomization (DR)~\cite{tobin2017domain,peng2018sim}: standard DR randomizes physical parameters to learn invariant features, while advanced variants further randomize simulator dynamics. Recent comprehensive studies have moved beyond isolated techniques to systematically identify the critical factors for zero-shot sim-to-real transfer. Chen et al.~\cite{chen2025simpleflight} found that accurate SysID is essential and indiscriminate DR can degrade performance, whereas selective DR, careful input-space design, action smoothness rewards, and large batch sizes are each indispensable. Their PPO-based framework, SimpleFlight, achieves over a $50\%$ reduction in trajectory tracking error on a real Crazyflie. In parallel, employing CTBR commands~\cite{mellinger2011minimum} grants the policy direct authority over attitude dynamics while delegating motor-level tracking to a robust low-level controller, and penalizing high-frequency oscillations ensures safe hardware deployment.

In summary, although differential games, MPC, and DRL have each advanced UAV PE, a unified solution that (i) directly outputs low-level actuator commands for peak agility, (ii) sustains tactical diversity through hierarchical adversarial training, and (iii) closes the sim-to-real gap without extensive fine-tuning, remains absent. To bridge these gaps, this paper proposes \textbf{AgilePE}, a tightly integrated system that combines an end-to-end CTBR control architecture with a bilateral adversarial training pipeline and a hardware-aligned deployment backend. The following sections detail each component and validate its effectiveness through large-scale simulation and physical experiments.

\section{AgilePE System Design}
\label{sec:system}

\begin{figure*}[t]
    \centering
    % \fbox{\parbox[c][8cm][c]{0.8\linewidth}{\centering \textbf{[Placeholder]}\\ System Overview Diagram}}
    \includegraphics[width=0.8\textwidth]{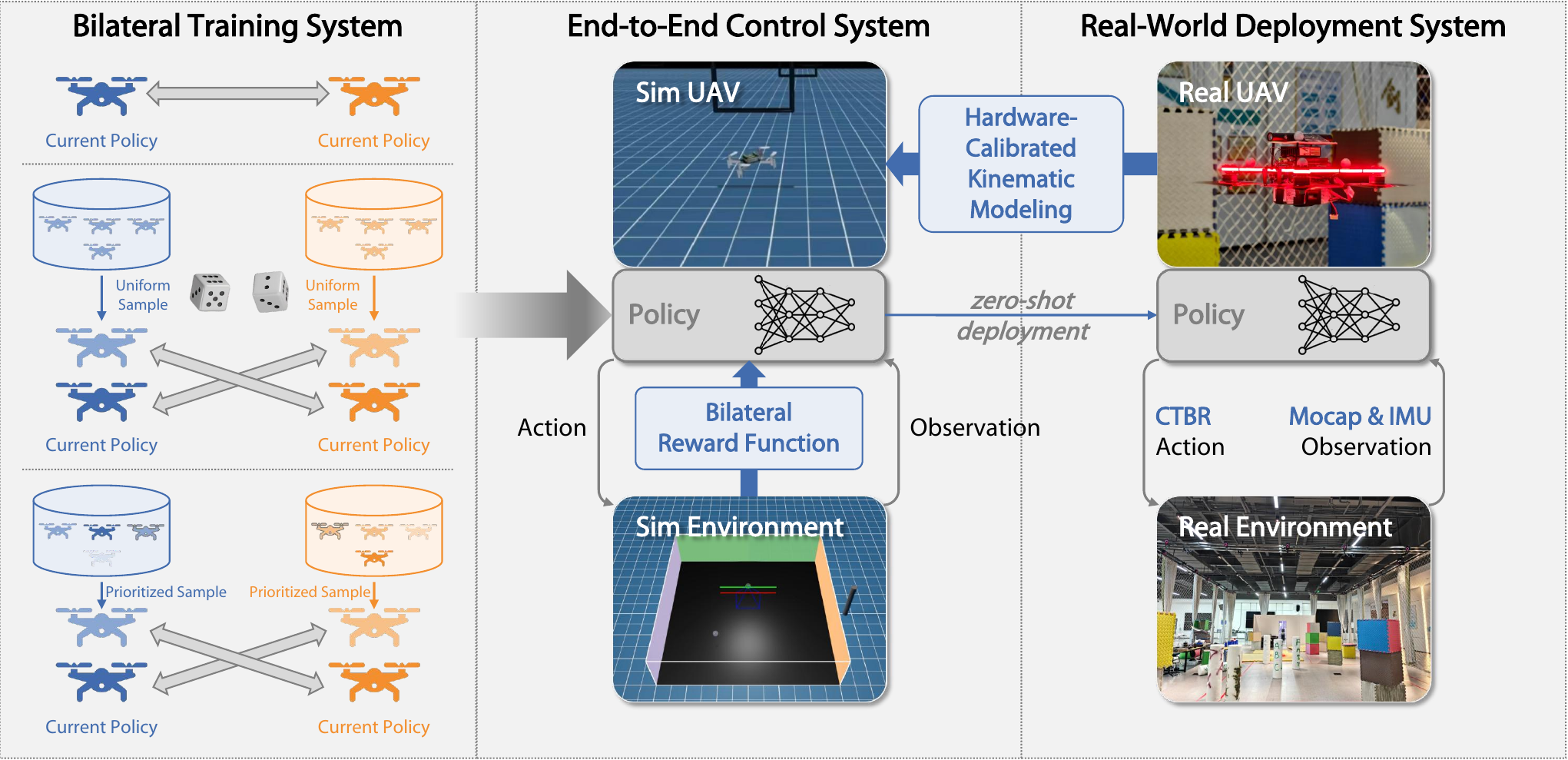}
    \caption{Overview of the AgilePE framework. The system comprises three tightly coupled subsystems: (i) the End-to-End Control Subsystem, (ii) the Bilateral Training Subsystem, and (iii) the Real-World Deployment Subsystem.}
    \label{fig:overview}
\end{figure*}

The AgilePE system is organized around three tightly coupled subsystems: (i) an \textit{End-to-End Control Subsystem} that defines the state-action interface, flight dynamics, and multi-objective reward architecture (Section \ref{subsec:control}); (ii) a \textit{Bilateral Training Subsystem} that governs adversarial policy evolution through hierarchical SP regimes (Section \ref{subsec:training}); and (iii) a \textit{Real-World Deployment Subsystem} that enables robust transfer to physical platforms (Section \ref{subsec:sim_backend}). The overall architecture and information flow among these subsystems are illustrated in Fig.~\ref{fig:overview}.

\subsection{Task Description}
\label{subsec:task}

\begin{figure}[t]
    \centering
    \includegraphics[width=0.8\linewidth]{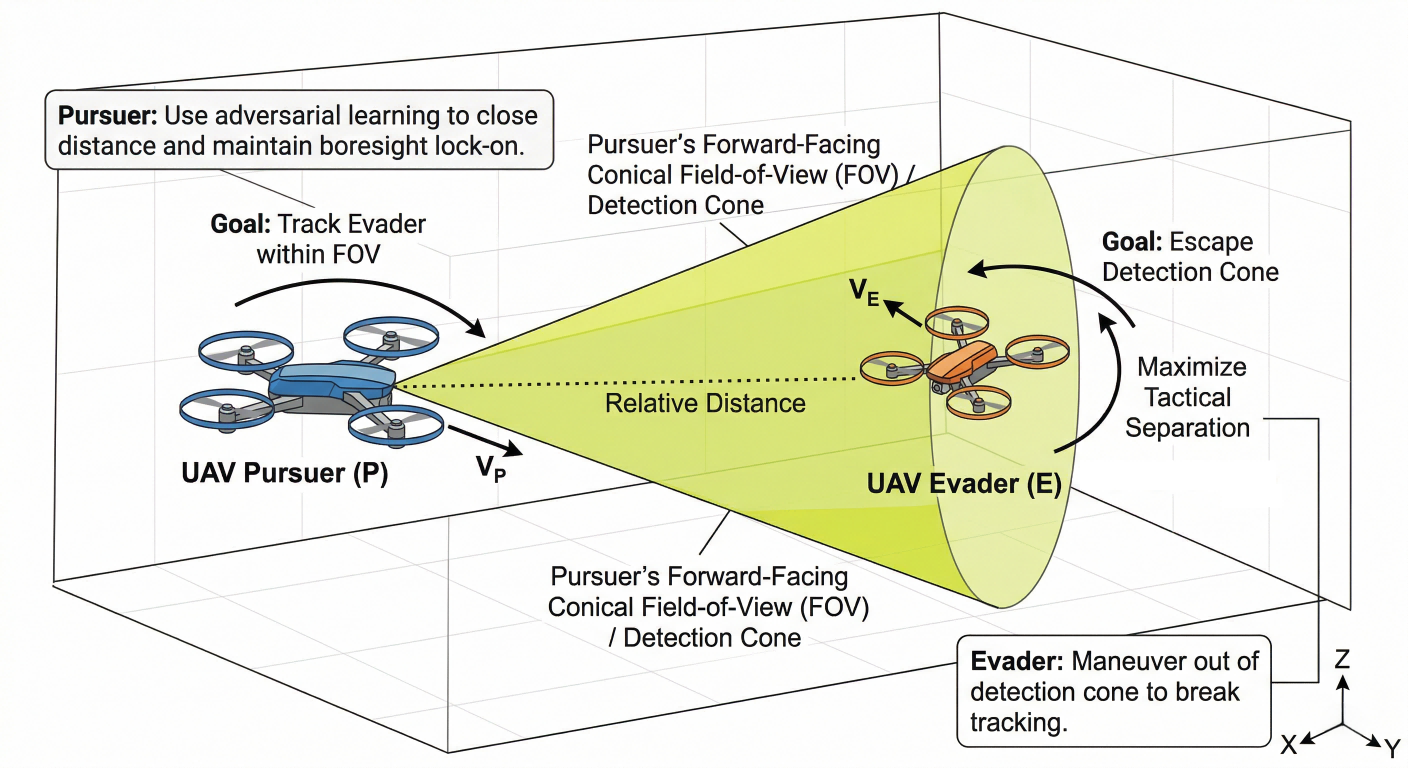}
    \caption{1v1 aerial pursuit-evasion problem formation. The pursuer (blue) must keep the evader (red) inside its conical FOV while closing the distance; the evader exploits obstacles and boundary geometry to break visual contact.}
    \label{fig:problem}
\end{figure}

We investigate 1-versus-1 (1v1) aerial PE formulated as a partially-observable zero-sum game within a bounded three-dimensional workspace, as shown in Fig.~\ref{fig:problem}~\cite{isaacs1965differential,vidal2002probabilistic}. The pursuer's objective is to minimize the relative distance to the evader while maintaining the target within its forward-facing conical Field-of-View (FOV). Conversely, the evader seeks to maneuver out of the pursuer's detection cone and maximize tactical separation.

The environment is built upon the OmniDrones framework~\cite{xu2024omnidrones} for large-scale parallel RL, with $N=2048$ independent arena instances running simultaneously. The workspace is a $6\,\mathrm{m} \times 6\,\mathrm{m} \times 3\,\mathrm{m}$ enclosed arena with invisible boundary walls. Both agents are modeled as 6-DOF rigid bodies driven by a CTBR interface~\cite{kaufmann2022benchmark}. %; full dynamics are deferred to Appendix~\ref{app:dynamics}.

Each agent's perception is restricted by a conical FOV and by occlusions from obstacles. The observation comprises the agent's own state (position, attitude, velocity, and past CTBR commands), a temporal window of the opponent's state (masked when outside FOV or occluded), and relative positions of additional agents if applicable. Visibility is determined by a frustum check combined with cylinder occlusion testing; missing entries are filled with a mask value. Both agents issue CTBR commands $\mathbf{u}=[T,\omega_x,\omega_y,\omega_z]^\top$, which are mapped to rotor speeds and propagated through the rigid-body dynamics. An episode terminates upon timeout, out-of-bounds, or collision when the inter-agent distance drops below a safety threshold.

\subsection{End-to-End Control Subsystem}
\label{subsec:control}

The control subsystem models each UAV as a 6-DOF rigid body with state $\mathbf{x}_t = [\mathbf{p}, \mathbf{q}, \mathbf{v}, \boldsymbol{\omega}]^T$ and actions $\mathbf{u} = [T_d, \omega_{x,d}, \omega_{y,d}, \omega_{z,d}]^T$ under the CTBR paradigm.

The reward function balances tactical objectives with physical safety:
\begin{equation}
    R = r_{\text{fov\_dist}} + \mathcal{P}_{\text{speed}} + \mathcal{P}_{\text{edge}} + \mathcal{P}_{\text{close}} + r_{\text{smooth}}
\end{equation}
To avoid prohibitively sparse win/loss signals, the FOV-distance term masks the inter-agent distance with a constant when the evader lies outside the detection cone:
\begin{equation}
    \tilde{d} =
    \begin{cases}
        d, & \text{if } \mathbb{I}_{\text{FOV}} = 1 \\
        D_{\max}, & \text{otherwise}
    \end{cases}
\end{equation}
The pursuer and evader then receive opposing rewards scaled by $\tilde{d}$, forming a zero-sum core that encourages persistent tracking under realistic sensing constraints.

Three safety regularizers ensure physical viability. The overspeed penalty $\mathcal{P}_{\text{speed}}$ applies an exponential cost when velocity exceeds the structural limit. The boundary penalty $\mathcal{P}_{\text{edge}}$ triggers a quadratic cost near the workspace boundary. The proximity penalty $\mathcal{P}_{\text{close}}$ discourages mid-air collisions below a safety distance. Finally, the smoothness reward $r_{\text{smooth}}$ penalizes high-frequency oscillations in consecutive CTBR commands, which is essential for stable zero-shot deployment on physical hardware.

\subsection{Bilateral Training Subsystem}
\label{subsec:training}

We build upon MAPPO~\cite{yu2021surprising} and instantiate three progressively more sophisticated adversarial training frameworks for the aerial PE task. Each framework maintains a population of historical policy checkpoints for both the pursuer and the evader, but differs in how opponents are sampled during co-evolution. In \textbf{naive SP}~\cite{heinrich2016deep}, the latest pursuer and evader policies face each other across all environments and update synchronously. This minimal ``arms race'' incurs no historical overhead and serves as an ablation baseline, yet it suffers from severe non-stationarity: each agent optimizes against a single, rapidly shifting opponent, leading to policy oscillation and catastrophic forgetting~\cite{mccloskey1989catastrophic}. \textbf{FSP}~\cite{heinrich2015fictitious} stabilizes training by splitting the environments into two halves: one half trains the pursuer against a uniformly sampled mixture of historical evader checkpoints, while the other half trains the evader against historical pursuer checkpoints. This diversification provides a stationary training target that approximates a Nash equilibrium~\cite{nash1950equilibrium}, which is critical because a single non-stationary opponent in high-dimensional continuous control induces prohibitive gradient variance. \textbf{PFSP}~\cite{vinyals2019grandmaster} further replaces uniform sampling with win-rate-based prioritization: the pursuer concentrates on evader history it historically defeats well, while the evader prioritizes pursuer opponents from which it most frequently escaped. This design improves sample efficiency by avoiding wasted interaction against already-mastered or overwhelmingly strong opponents, thereby accelerating the emergence of high-level maneuvers within reasonable wall-clock time.

% Even with FSP or PFSP, random initialization in a sparse-reward competitive game rarely yields meaningful gradients. We therefore precede SP with a warm-up phase in which each agent is bootstrapped against fixed rule-based opponents until convergence, with roles alternated across iterations. To further stabilize early learning, we introduce \textbf{Online Curriculum Learning (OCL)} to mitigate the information asymmetry inherent in aerial PE. Both the pursuer and the evader are initially granted an expanded perceptual field beyond their nominal conical FOV; as training progresses, this auxiliary awareness is progressively annealed until both agents operate under the realistic restricted sensing conditions. By starting from an information-rich regime, OCL enables rapid bootstrapping of basic tactical behaviors; the subsequent withdrawal of perceptual scaffolding then forces each agent to generalize to the true partially observable setting. Detailed curriculum parameters are provided in Appendix~\ref{app:curriculum}.

\subsection{Real-World Deployment Subsystem}
\label{subsec:sim_backend}

The deployment backend centers on a hardware-calibrated kinematic integration scheme that replaces uncertain physics-engine contact solvers with a deterministic model. Unlike standard force-torque-based simulations in Isaac Sim, this approach directly integrates the UAV's pose and velocity using the CTBR control interface, providing deterministic and numerically stable state evolution during aggressive maneuvers. The translational acceleration is governed by the thrust vector and gravity:
\begin{equation}
    \dot{\mathbf{v}} = \mathbf{R}_B^W \cdot [0, 0, T]^T - \mathbf{g}
\end{equation}
where $\mathbf{R}_B^W$ is the rotation matrix from the body frame to the world frame and $\mathbf{g} = [0, 0, 9.81]^T$. To maintain numerical stability during high-speed pursuit, we employ a \textbf{fourth-order Runge-Kutta (RK4)} scheme with $10$ integration sub-steps per control cycle ($\Delta t = 0.016$ s, $dt = 0.0016$ s).

A critical component of the Sim-to-Real gap~\cite{tobin2017domain,rusu2016sim,peng2018sim} is the discrepancy in actuator response. We model this through a \textit{ResponseTool} that combines first-order latency with stochastic noise. Thrust and body-rate commands are processed through a sliding-window delay characterized by nominal response times ($\tau_{\text{thrust}} = 45$ ms, $\tau_{\omega} = 30$ ms), while multiplicative noise ($\pm 10\%$ on thrust, $\pm 30\%$ on body rates) is injected during training to enhance robustness against hardware variance. This explicit actuator emulation is necessary because policies trained in idealized simulations often issue high-frequency commands that destabilize physical airframes; by baking hardware response characteristics into the training loop, the policy learns to compensate for latency rather than being surprised by it at deployment.

The resulting end-to-end deployment pipeline generates CTBR commands through the policy network,  emulates actuator latency and noise, integrates the resulting kinematic state using RK4 over ten sub-steps per control cycle, and updates poses directly within Isaac Sim while bypassing the internal physics solver. Key simulation and domain-randomization parameters are summarized in Table~\ref{tab:sim_params}.

\begin{table}[htbp]
\centering
\caption{Key Parameters of the Kinematic Simulation and Sim2Real Configuration.}
\label{tab:sim_params}
\begin{tabular}{@{}lll@{}}
\toprule
\textbf{Parameter} & \textbf{Value} & \textbf{Description} \\ \midrule
\textit{Integration} & & \\
Outer Step ($\Delta t$) & $0.016$ s & Policy frequency (60Hz) \\
Inner Step ($dt$) & $0.0016$ s & RK4 integration step \\
RK4 Iterations & $10$ & Sub-steps per control cycle \\ \midrule
\textit{Response Latency} & & \\
Thrust Delay ($\tau_{\text{thrust}}$) & $45$ ms & Nominal thrust response time \\
Rate Delay ($\tau_{\omega}$) & $30$ ms & Nominal body rate response time \\
Smoothing Window ($\lambda$) & $6.4$ ms & Response smoothing parameter \\ \midrule
\textit{Domain Randomization} & & \\
Thrust Noise & $\pm 10\%$ & Amplitude gain error \\
Body Rate Noise & $\pm 30\%$ & Amplitude gain error \\
Delay Jitter ($\tau_{\text{thrust}}$) & $\pm 30\%$ & Temporal latency fluctuation \\
Delay Jitter ($\tau_{\omega}$) & $\pm 20\%$ & Temporal latency fluctuation \\ \bottomrule
\end{tabular}
\end{table}

\section{Simulation Experiments and Results}
\label{sec:training_pipeline}

We evaluate the efficacy of the AgilePE system within OmniDrones, a high-fidelity simulation platform designed for UAV reinforcement learning. All policies are trained using the systematic pipeline described in Section \ref{subsec:training}, which comprises a warm-up phase, multi-dimensional curriculum learning, and hierarchical adversarial training frameworks. This section reports the quantitative performance of the three adversarial regimes and their cross-evaluation against scripted baselines.

\subsection{Bilateral Adversarial Framework Evaluation}

To validate the design of the bilateral training subsystem, we instantiate and compare three hierarchical adversarial frameworks: naive SP, FSP, and PFSP. Each framework represents a progressively more sophisticated approach to maintaining tactical diversity and driving policy evolution.

\subsubsection{Naive self-play}

\begin{figure}[t]
    \centering
    \includegraphics[width=0.8\linewidth]{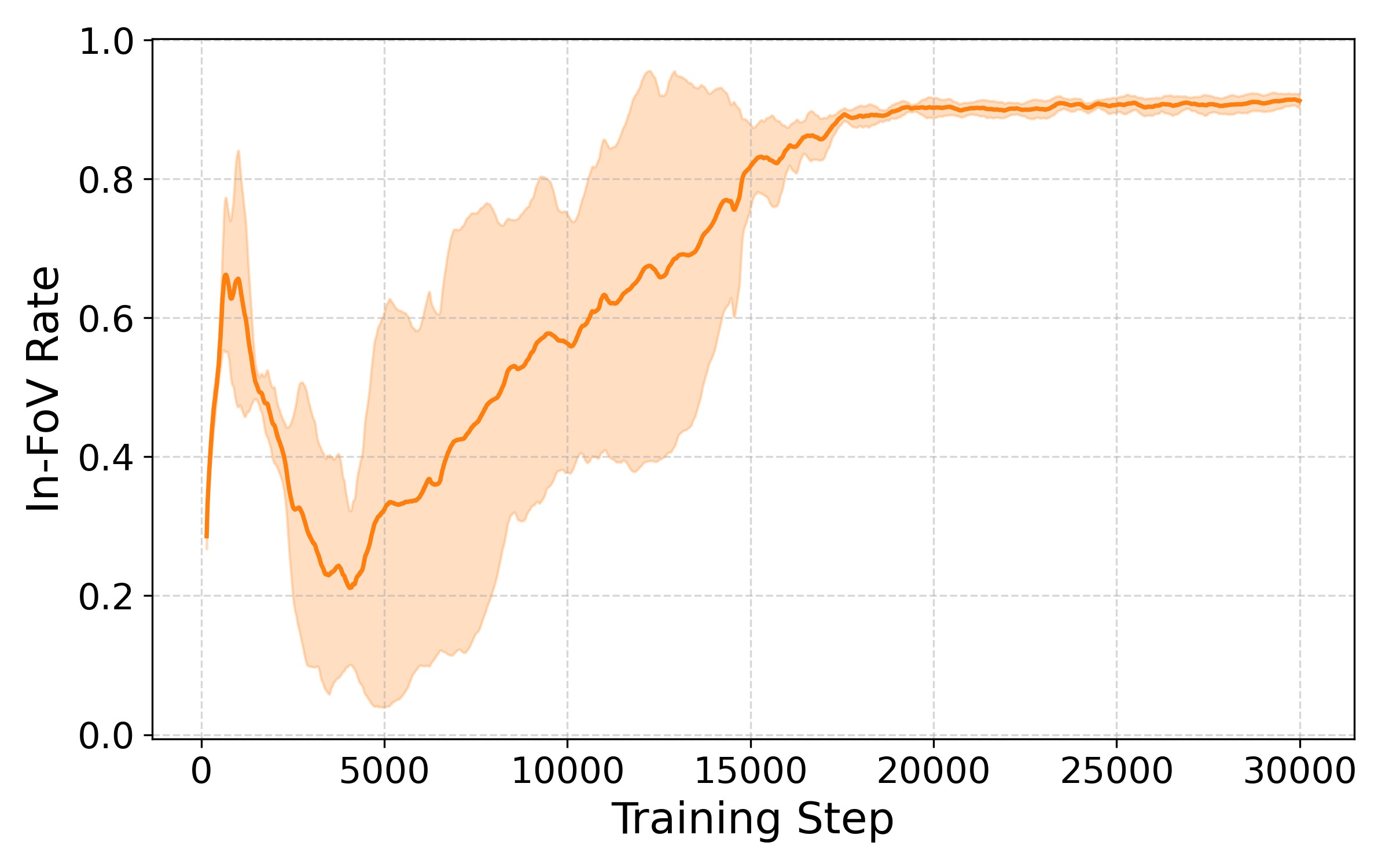}
    \caption{The evolution curve of the in-FOV rate for naive SP training}
    \label{fig:self_play}
\end{figure}

To evaluate the robustness and reproducibility of the naive SP framework, we conducted training across multiple independent trials using different random seeds. Figure \ref{fig:self_play} illustrates the evolution of the in-FOV rate—a key performance indicator for pursuit success—as a function of training iterations. 

The evolution of the in-FOV rate across different seeds reveals two distinct phases of adversarial co-evolution. In the early training stage, the significant oscillations in performance suggest a rich exploration of tactical maneuvers, where both the pursuer and evader iteratively develop counter-strategies, leading to diversified adversarial behaviors. Detailed analysis of these emergent behaviors is further elaborated in Section \ref{sec:emergent}.

However, in the late training stage, the in-FOV rate across all seeds converges to a disproportionately high value (exceeding $0.9$). In a bilateral training context, this phenomenon indicates a unilateral strategy collapse of the evader. Rather than reaching a robust Nash Equilibrium, the SP process tends to stagnate into a sub-optimal equilibrium, where the evader fails to discover effective defensive maneuvers against the pursuer's stabilized policy. This lack of sustained competitive pressure in the later stages highlights the inherent limitation of naive SP in maintaining strategy diversity.

\subsubsection{Fictitious self-play (FSP)}

\begin{figure}[t]
    \centering
    % 第一张图：FSP
    \begin{subfigure}[b]{0.49\linewidth}
        \centering
        \includegraphics[width=\textwidth]{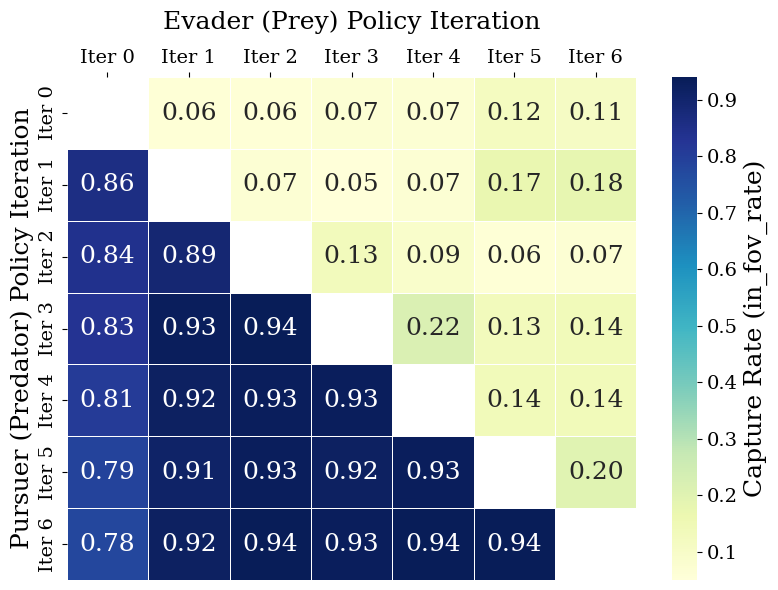}
        \caption{FSP}
        \label{fig:payoff_fsp}
    \end{subfigure}
    \hfill 
    % 第二张图：PFSP
    \begin{subfigure}[b]{0.49\linewidth}
        \centering
        \includegraphics[width=\textwidth]{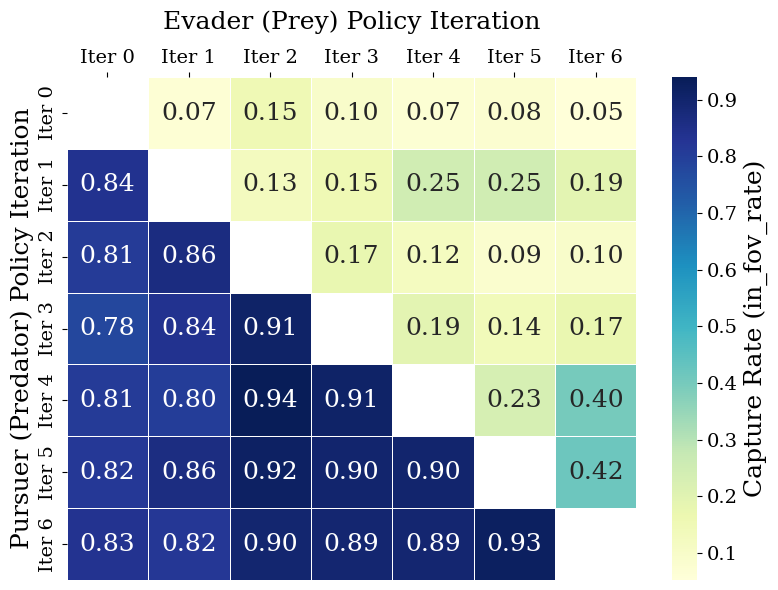}
        \caption{PFSP}
        \label{fig:payoff_pfsp}
    \end{subfigure}
    
    \caption{Cross-evaluation payoff matrices comparing the policy evolution of FSP and PFSP. Each cell $(i, j)$ denotes the mean in-FOV rate of Predator $i$ against Prey $j$. Note the higher saturation in the lower-left triangle of (b), indicating more robust strategy dominance.}
    \label{fig:payoff_comparison}
\end{figure}

To overcome the strategy collapse observed in naive SP, we implement the FSP regime. In this framework, we maintain a historical strategy pool $\mathcal{H} = \{\pi_0, \pi_1, \dots, \pi_{k-1}\}$ for both the pursuer and the evader. During the $k$-th training iteration, the active agent is trained against a target policy sampled uniformly from the opponent's strategy pool, excluding the current version. To ensure continuous tactical evolution, each training round is warm-started using the network weights from the previous iteration. Upon completion of a round, the optimized policy is archived into the pool, progressively expanding the diversity of the training distribution.

The effectiveness of FSP is evaluated through a cross-evaluation payoff matrix (as shown in Figure \ref{fig:payoff_fsp}), where the in-FOV rate is measured for all combinations of archived pursuer and evader policies. Unlike the stagnation seen in naive SP, the FSP matrix exhibits a clear evolutionary progression. Later iterations of the pursuer (e.g., Predator 6) demonstrate superior tracking performance against a broader range of historical evaders, while later evaders (e.g., Prey 6) exhibit enhanced survival capabilities against earlier predators.

We further validate the learned policies against fixed scripted opponents (rule-based agents). As shown in the comparative results of Table \ref{tab:sp_script_results} and Table \ref{tab:fsp_script_results}, the FSP-trained evader achieves a significant improvement in survival rate, with the in-FOV rate against a scripted predator increasing from $0.28$ (iteration 0) to $0.81$ (iteration 4). In contrast, the naive SP policy tends to maintain a mediocre performance (approx. $0.5$) due to its inability to retain counter-strategies against diverse behaviors. These results demonstrate that FSP effectively suppresses policy forgetting and guides the agents toward a more robust and generalized adversarial strategy.

\subsubsection{Prioritized fictitious self-play (PFSP)}

To further accelerate the emergence of high-level tactical behaviors, we extend FSP into a PFSP framework. The primary innovation lies in the weighted sampling mechanism used to select opponents from the strategy pool $\mathcal{H}$. Instead of uniform selection, the sampling probability for each historical strategy $\pi_i \in \mathcal{H}$ is proportional to the current agent's failure rate against it. Specifically, we use the initial win rate (or capture rate) of the current policy against each candidate in the pool to calculate a priority score, ensuring that the training process focuses on the most challenging "bottleneck" opponents. Similar to FSP, PFSP maintains an incremental training regime where each round inherits weights from the previous one, and the resulting optimized policy is added to the strategy pool upon convergence.

As shown in the cross-evaluation matrix (Figure \ref{fig:payoff_pfsp}), the PFSP-trained Predator (Iter 6) achieves a consistent in-FOV rate of over $0.82$ against all historical versions of the Prey. While FSP provides a stable foundation for preventing policy collapse through a diversified training distribution, PFSP acts as an automated curriculum, pushing the agents toward the theoretical limits of their tactical capabilities. As shown in Table \ref{tab:pfsp_script_results}, the experimental results reveal that PFSP achieves a superior capture rate of 0.88 against scripted baselines. However, this intensified focus on 'hard' RL adversaries induces a tactical specialization effect, where the agents become slightly more vulnerable to simplistic scripted patterns in exchange for dominance over sophisticated, evolving opponents. This trade-off underscores the effectiveness of prioritized sampling in accelerating the emergence of high-tier maneuvers.

\begin{table}[t]
\centering
\caption{Quantitative evaluation of different policies against scripted baselines (In-FOV Rate).}
\label{tab:combined_script_results}

% ------------------ 子表 1 ------------------
\begin{subtable}{\linewidth}
\centering
\caption{Performance of naive SP policies.}
\label{tab:sp_script_results}
\small
\setlength{\tabcolsep}{3pt}
\begin{tabular}{@{}lcccccc@{}}
\toprule
\textbf{Training Steps} & \textbf{1k} & \textbf{3k} & \textbf{5k} & \textbf{10k} & \textbf{15k} & \textbf{30k} \\ \midrule
SP Pred. vs. Scr. Prey & 0.49 & 0.28 & 0.32 & 0.45 & 0.58 & 0.42 \\
SP Prey vs. Scr. Pred. & 0.16 & 0.11 & 0.06 & 0.06 & 0.15 & 0.50 \\ \bottomrule
\end{tabular}
\end{subtable}

\vspace{1em} % 在子表之间留出一点垂直间距

% ------------------ 子表 2 ------------------
\begin{subtable}{\linewidth}
\centering
\caption{Evaluation of FSP policies.}
\label{tab:fsp_script_results}
\small
\setlength{\tabcolsep}{3pt}
\begin{tabular}{@{}lcccccccc@{}}
\toprule
\textbf{Iter.} & \textbf{Scr.} & \textbf{0} & \textbf{1} & \textbf{2} & \textbf{3} & \textbf{4} & \textbf{5} & \textbf{6} \\ \midrule
Pred. vs. Prey $k$ & 0.16 & 0.12 & 0.06 & 0.05 & 0.05 & 0.03 & 0.03 & 0.05 \\
Prey vs. Pred. $k$   & 0.16 & 0.28 & 0.26 & 0.30 & 0.66 & 0.82 & 0.78 & 0.81 \\ \bottomrule
\end{tabular}
\end{subtable}

\vspace{1em} % 在子表之间留出一点垂直间距

% ------------------ 子表 3 ------------------
\begin{subtable}{\linewidth}
\centering
\caption{Evaluation of PFSP policies.}
\label{tab:pfsp_script_results}
\small
\setlength{\tabcolsep}{3pt}
\begin{tabular}{@{}lcccccccc@{}}
\toprule
\textbf{Iter.} & \textbf{Scr.} & \textbf{0} & \textbf{1} & \textbf{2} & \textbf{3} & \textbf{4} & \textbf{5} & \textbf{6} \\ \midrule
Pred. vs. Prey $k$ & 0.16 & 0.08 & 0.07 & 0.34 & 0.58 & 0.58 & 0.58 & 0.30 \\
Prey vs. Pred. $k$   & 0.16 & 0.53 & 0.64 & 0.69 & 0.75 & 0.84 & 0.88 & 0.88 \\ \bottomrule
\end{tabular}
\end{subtable}

\end{table}

\section{Emergent Behavior Analysis}
\label{sec:emergent}

\begin{figure*}[htbp]
    \centering
    \begin{subfigure}{\textwidth}
        \centering
        \includegraphics[width=\textwidth]{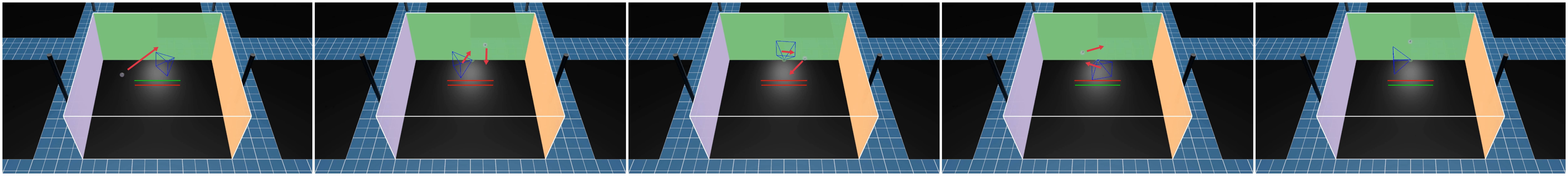}
        \caption{\textbf{Evader Tactic (Prey vs. Scripted Pursuer):} The evader executes a rapid flanking maneuver to the pursuer's rear, followed by circling to exploit rotational latency.}
        \label{fig:evader_script}
    \end{subfigure}\\[2ex]
    \begin{subfigure}{\textwidth}
        \centering
        \includegraphics[width=\textwidth]{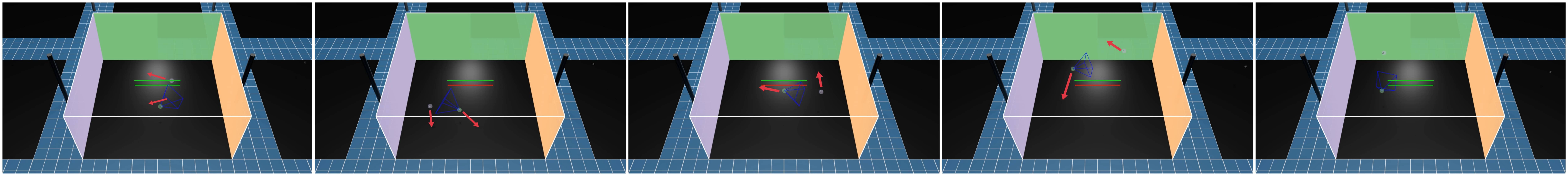}
        \caption{\textbf{Pursuer Tactic (Predator vs. Spinning Evader):} The pursuer maintains a tactical standoff distance to reduce relative angular velocity and ensure stable tracking.}
        \label{fig:pursuer_spinning}
    \end{subfigure}
    \caption{Case studies of emergent interpretable behaviors against diverse opponent types, where red arrows denote the instantaneous velocity vectors of the UAVs.}
    \label{fig:emergent_behaviors_case}
\end{figure*}

To assess the tactical depth of the learned policies, we investigate their performance against specific scripted and heterogeneous opponents. The emergence of interpretable maneuvers highlights the robustness and generalization of the adversarial framework.

\paragraph{Tactical Flanking and Latency Exploitation}
When engaged with a scripted pursuer utilizing a rigid direct-tracking heuristic, the SP-trained evader exhibits a sophisticated \textbf{"dash-and-flank"} maneuver (Fig. \ref{fig:evader_script}). The evader initially executes a high-speed dash to the pursuer's rear hemisphere, a tactical blind spot. Upon reaching this position, it momentarily decelerates, effectively "waiting" for the pursuer to initiate a re-orientation turn. Subsequently, the evader maintains a circling trajectory, precisely exploiting the pursuer's rotational latency to remain outside the detection cone. This behavior demonstrates the agent's ability to discover and exploit the structural vulnerabilities of rule-based adversaries.

\paragraph{Distance Management and Precise Tracking}
In scenarios involving a heterogeneous "fast-spinning" evader, the SP-trained pursuer develops a conservative yet effective tracking strategy (Fig. \ref{fig:pursuer_spinning}). Rather than aggressively closing the range, which often results in overshooting or tracking instability against spinning targets, the pursuer proactively maintains a \textbf{strategic standoff distance}. From this optimal range, the pursuer employs small-amplitude angular adjustments to sustain boresight lock-on. This tactic minimizes the relative angular velocity required for tracking, ensuring robust engagement against anomalous, high-frequency evasive patterns.

\section{Real-World Validation}
\label{sec:sim2real}

\begin{figure*}[t]
    \centering
    \begin{subfigure}{\textwidth}
        \centering
        \includegraphics[width=\textwidth]{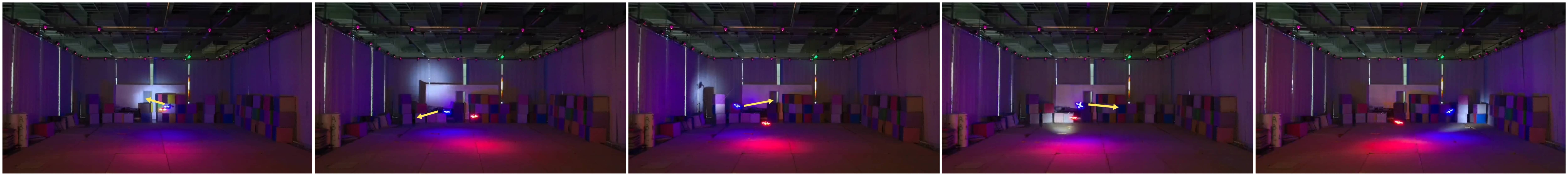}
        \caption{\textbf{Dash-and-Flank Maneuver:} The evader dashes toward the side boundary and rapidly flanks behind the pursuer to break visual lock.}
        \label{fig:real_1}
    \end{subfigure}\\[1.5ex]
    \begin{subfigure}{\textwidth}
        \centering
        \includegraphics[width=\textwidth]{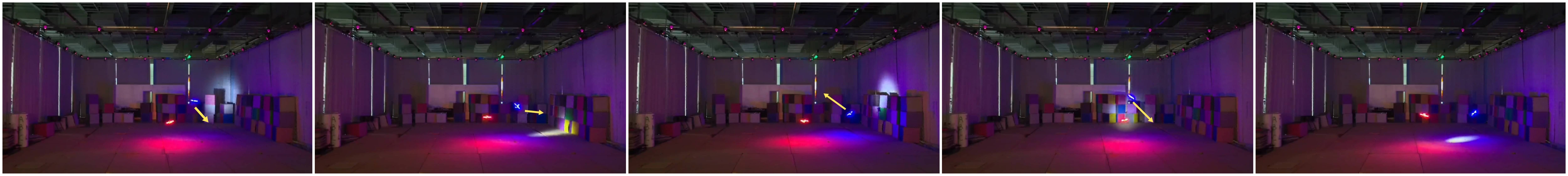}
        \caption{\textbf{Lateral Oscillation:} The evader performs rapid left-right oscillations to exploit the pursuer's tracking latency and escape the field of view.}
        \label{fig:real_2}
    \end{subfigure}
    \caption{Real-world validation snapshots illustrating emergent evasion tactics executed by the our quadrotors. The blue drone represents the evader and the red drone represents the pursuer. Yellow arrows indicate the instantaneous velocity direction of the evader.}
    \label{fig:real_world_validation}
\end{figure*}

This section presents the physical validation of the AgilePE system on real quadrotor hardware (see Fig.~\ref{fig:teaser}). The deployment leverages the hardware-aligned simulation backend detailed in Section \ref{subsec:sim_backend}, which calibrates actuator latencies, injects domain randomization, and enforces control smoothness to enable zero-shot transfer.

The trained policy runs onboard at 60Hz on a Jetson Orin NX computer, generating CTBR commands that are executed by an onboard PX4 Autopilot flight controller. State estimation is provided by an external motion capture system streaming pose measurements at 120Hz, which are fused with onboard IMU data to generate high-frequency state feedback. The flight controller is carefully tuned to respond to commands with a bodyrate delay of less than 20ms. All inference is performed on the Jetson Orin NX, enabling fully autonomous, GPU-accelerated policy execution without offboard computation. Preliminary hardware tests confirm that the policy maintains stable flight characteristics. Furthermore, we conducted aerial combat validation on physical platforms to verify the emergence of adversarial maneuvers learned in simulation.

\paragraph{Dash-and-Flank Maneuver}
As illustrated in Fig.~\ref{fig:real_1}, the evader first executes a high-speed dash toward one side of the arena. Upon reaching the lateral boundary, it rapidly reverses direction and maneuvers behind the pursuer, effectively breaking the pursuer's line-of-sight lock by exploiting the pursuer's turning latency. This behavior mirrors the flanking tactics observed in simulation, demonstrating successful zero-shot transfer of the learned policy to physical hardware.

\paragraph{Lateral Oscillation for Visual Break}
Fig.~\ref{fig:real_2} presents a second emergent tactic where the evader employs rapid lateral oscillations to escape the pursuer's field of view. By alternating abrupt left-right accelerations, the evader introduces high relative angular velocity that complicates the pursuer's tracking efforts, leading to repeated loss of visual lock.

\section{Conclusion and Future Work}
\label{sec:conclusion}

We presented \textbf{AgilePE}, an integrated system for autonomous UAV pursuit-evasion that couples end-to-end CTBR control, bilateral adversarial training (SP, FSP, PFSP), and a hardware-aligned sim-to-real pipeline. Our experiments demonstrate that PFSP sustains tactical diversity and drives robust policy equilibria, while the learned policies exhibit emergent maneuvers such as flanking and distance management without hand-crafted reward shaping. Zero-shot deployment on physical quadrotors validates the practical efficacy of the proposed framework.

Despite these results, several limitations remain. First, experiments were restricted to 1v1 engagements in obstacle-free environments; scaling to multi-agent coordination and cluttered terrain is an important next step. Second, the agents relied on state-based observations; integrating vision-based perception would enhance autonomy in unstructured environments where precise state estimation is unavailable.

We believe that the methodologies developed here lay a solid foundation for more complex, multi-agent autonomous systems in dynamic and contested environments.

\bibliographystyle{IEEEtran}
\bibliography{references}

% \newpage
% \input{tex/A_appendix}
% {\input{tex/A_appendix}}

\end{document}